\documentclass[10pt,journal, twoside]{IEEEtran}
\usepackage{amsmath,amsfonts}
\usepackage{algorithm}         
\usepackage[noend]{algpseudocode}
\usepackage{array}
\usepackage[caption=false,font=normalsize,labelfont=sf,textfont=sf]{subfig}
\usepackage{textcomp}
\usepackage{stfloats}
\usepackage{url}
\usepackage{verbatim}
\usepackage{graphicx}
\def\BibTeX{{\rm B\kern-.05em{\sc i\kern-.025em b}\kern-.08em
    T\kern-.1667em\lower.7ex\hbox{E}\kern-.125emX}}
\usepackage{balance}

\usepackage{booktabs}
\usepackage{subfig}
\usepackage[dvipsnames]{xcolor}
\usepackage{eso-pic}

\newboolean{showcomments}
\setboolean{showcomments}{false}
\ifthenelse{\boolean{showcomments}}
{ \newcommand{\mynote}[3]{
		\fbox{\bfseries\sffamily\scriptsize#1}
		{\small$\blacktriangleright$\textsf{\emph{\color{#3}{#2}}}$\blacktriangleleft$}}}
{ \newcommand{\mynote}[3]{}}
\newcommand{\shrink}[1]{}
\definecolor{pink}{rgb}{1,0.2,0.7}
\definecolor{purple}{rgb}{0.7,0,0.9}

\newcommand{\jb}[1]{\mynote{Jalil}{#1}{purple}}

\begin{document}
\title{RETENTION: \underline{R}esource-\underline{E}fficient \underline{T}ree-Based \underline{En}semble Model Accelera\underline{tion} with Content-Addressable Memory}

\author{
\thanks{Manuscript received 6 June 2025; revised 4 February 2026 and 20 April 2026; accepted 22 April 2026. Date of publication 7 May 2026.}
\thanks{This work was supported in part by National Science and Technology Council under grant nos. 114-2927-I-002-525, 115-2927-I-002-519, 114-2223-E-002-011, 114-2221-E-002-219-MY3, 114-2221-E-002-222-MY3, 115-2923-E-002-008-MY3, 114-2221-E-002-106-MY3, and the financial supports from the Featured Area Research Center Program within the framework of the Higher Education Sprout Project by the Ministry of Education (114L900903, 115L900903). \textit{(Corresponding author: Yuan-Hao Chang)}}

Yi-Chun Liao,
\thanks{Yi-Chun Liao is with the Department of Computer Science and Information Engineering, National Taiwan University, Taipei 10617, Taiwan. E-mail: ycliao718@gmail.com.}
Chieh-Lin Tsai,
\thanks{Chieh-Lin Tsai is with the Department of Computer Science and Information Engineering, National Taiwan University, Taipei 10617, Taiwan. E-mail: d09922013@csie.ntu.edu.tw.}
Yuan-Hao Chang, ~\IEEEmembership{Fellow, ~IEEE,}
\thanks{Yuan-Hao Chang is with the Department of Computer Science and Information Engineering, National Taiwan University, Taipei 10617, Taiwan. E-mail: johnson@csie.ntu.edu.tw.}
Camélia Slimani, \\
\thanks{Camélia Slimani is with IRIT, Université de Toulouse, Toulouse INP–UT3, CNRS, 31062 Toulouse, France. E-mail: camelia.slimani@toulouse-inp.fr.}
Jalil Boukhobza,~\IEEEmembership{Senior Member,~IEEE,} 
\thanks{Jalil Boukhobza is with Lab-STICC, CNRS UMR 6285 , ENSTA, Institut Polytechnique de Paris, 29806 Brest, France. E-mail: jalil.boukhobza@ensta.fr.}
and Tei-Wei Kuo, ~\IEEEmembership{Fellow, ~IEEE}
\thanks{Tei-Wei Kuo is with the Department of Computer Science and Information Engineering, National Taiwan University, Taipei 10617, Taiwan. He is also with Delta Electronics, Inc., Taipei 114501, Taiwan. E-mail: ktw@csie.ntu.edu.tw.}
\thanks{RETENTION is available at https://github.com/datou0718/RETENTION}
}

\markboth{IEEE Transactions on Computer-Aided Design of Integrated Circuits and Systems,~Vol.~45, No.~11, November~2026}
{\NoCaseChange{Yi-Chun Liao \textit{et al.}}: RETENTION: \underline{R}esource-\underline{E}fficient \underline{T}ree-Based \underline{En}semble Model Accelera\underline{tion} with content-addressable memory}

\maketitle

\AddToShipoutPictureFG*{%
  \AtPageLowerLeft{%
    \put(\LenToUnit{\dimexpr 1in+\oddsidemargin\relax},
         \LenToUnit{5mm}){%
      \parbox[b]{\textwidth}{%
        \fontsize{7.5}{8.5}\selectfont
        \raggedright
        \copyright~2026 IEEE. Personal use of this material is
        permitted. Permission from IEEE must be obtained for all
        other uses, in any current or future media, including
        reprinting/republishing this material for advertising or
        promotional purposes, creating new collective works, for
        resale or redistribution to servers or lists, or reuse of
        any copyrighted component of this work in other works.
      }%
    }%
  }%
}

\begin{abstract}

Although deep learning has demonstrated remarkable capability in learning from unstructured data, modern tree-based ensemble models remain superior in extracting relevant information and learning from structured datasets. While several efforts have been made to accelerate tree-based models, the inherent characteristics of the models pose significant challenges for conventional accelerators. Recent research leveraging content-addressable memory (CAM) offers a promising solution for accelerating tree-based models, yet existing designs suffer from excessive memory consumption and low utilization. This work addresses these challenges by introducing RETENTION, an end-to-end framework that significantly reduces CAM capacity requirement for tree-based model inference. We propose an iterative pruning algorithm with a novel pruning criterion tailored for bagging-based models (e.g., Random Forest), which minimizes model complexity while ensuring controlled accuracy degradation. Additionally, we present a tree mapping scheme that incorporates two innovative data placement strategies to alleviate the memory redundancy caused by the widespread use of don't care states in CAM. Experimental results show that implementing the tree mapping scheme alone reduces CAM capacity requirement by $1.46\times$ to $21.30 \times$, while the full RETENTION framework achieves $4.35\times$ to $207.12\times$ reduction with less than 3\% accuracy loss. These results demonstrate that RETENTION is highly effective in minimizing CAM resource demand, providing a resource-efficient direction for tree-based model acceleration.

\end{abstract}


\section{Introduction}

Structured (i.e., tabular) data is one of the most prevalent formats in data science. It is typically represented as a matrix, where each row corresponds to an instance and all instances share the same set of features across columns. This format is widely used in fields such as finance, medicine, and scientific research, as its structured nature enables efficient processing. For example, sensors generate data in tabular format to support real-time analysis for AI-driven closed-loop control. Despite significant advancements in deep learning for unstructured data (e.g., text, images, and speech), studies \cite{DLisNotAllYouNeed, TreeOutperformDL} have shown that modern tree-based ensemble models consistently outperform deep learning in several tasks. In fact, tree-based models remain the state-of-the-art for classification and regression tasks involving medium-sized structured datasets \cite{TreeOutperformDL}, and are often favored over deep learning due to their high interpretability \cite{Explainability}, particularly in sensitive applications where understanding model decisions are critical \cite{DARPA}. Tree-based models are widely applied in domains such as scientific research \cite{Science}, credit card fraud detection \cite{CreditCard}, machinery fault diagnosis \cite{FaultDiagnosis}, etc. A recent survey \cite{Kaggle} found that over 74\% of data scientists prefer tree-based models, whereas fewer than 40\% opt for neural networks, underscoring their continued importance. 

However, despite their widespread adoption and effectiveness, tree-based models have received less attention in recent years, and the inefficiency of tree-based model inference remains a critical yet unresolved challenge. Since inference requires multiple tree traversals, and the number of possible paths grows exponentially with tree depth, predicting and prefetching data becomes challenging. This may lead to relatively high inference latency, which is particularly unfriendly for real-time applications. The issue is further exacerbated as modern tree-based models (e.g., XGBoost \cite{XGBoost}) can contain thousands of trees, and are often deployed in resource-constrained environments \cite{EdgeML}. 

Although various accelerators have been proposed in recent years \cite{Accelerator1, Accelerator2}, researchers \cite{ConventionalDoesntWork} have found that conventional accelerators, such as multi-core CPUs, GP-GPUs, and FPGAs, offer limited effectiveness for tree-based model acceleration due to the non-deterministic memory access patterns and irregular tree structures. One promising approach to improving tree-based model inference is leveraging in-memory computing (IMC), which integrates data storage and processing within the same location, thereby eliminating latency and energy costs associated with data access and transfer. While conventional SRAM-based IMC accelerators can effectively accelerate model inference, they come at the cost of high energy consumption and extensive area overhead \cite{SRAM}, making it impractical for resource-constrained environments. 

Recently, there has been significant interest in developing emerging non-volatile-memory-based (NVM-based) IMC accelerators \cite{ISAAC, PRIME, NVMCIM}, as the characteristics of emerging NVM (e.g., high density, low power consumption, and low cost) make it well-suited for resource-constrained environments. Among potential candidates, non-volatile ternary content-addressable memory (nvTCAM) \cite{nvTCAM1, nvTCAM2, ReRAMTCAM} exhibits the best fit for accelerating tree-based models due to its capability to perform sequence matching with extremely high parallelism, energy efficiency, and reliability. Since every root-to-leaf path within tree-based models can be encoded into a binary sequence, nvTCAM can traverse all the paths in one shot, enabling unprecedented acceleration for model inference. Nevertheless, to support in-memory search, data must be organized in a specialized format (refer to Section~\ref{bg:CAM} for detailed explanation). Although nvTCAM effectively accelerates tree-based model inference in a cost- and energy-efficient manner, a considerable portion of memory cells is allocated for format alignment rather than storing actual model data. This leads to excessive memory consumption with substantial redundancy, which is highly inefficient and requires further optimization. 

To achieve resource-efficient acceleration, the enormous CAM capacity requirement with considerable redundancy is the major obstacle that must be addressed before practical implementation. Since model complexity is highly correlated with memory consumption, pruning models can significantly mitigate this issue. Modern tree-based models often employ pruning algorithms, such as limiting maximum depth or minimum impurity decrease \cite{Sklearn}, in order to reduce overfitting and simplify model structure. While these techniques are well-suited for boosting-based ensemble models (e.g., XGBoost), where each tree in the ensemble is trained to correct the errors of the previous ones, applying them to bagging-based ensemble models (e.g., Random Forest \cite{RandomForest}) can lead to severe accuracy degradation due to the independent training of each tree (refer to Section~\ref{method:pruning} for detailed explanation). On the other hand, memory redundancy arises from structuring data for in-memory search, suggesting that a tailored data placement strategy could further reduce CAM capacity requirement. While several studies \cite{DT2CAM, TreeBasedACAM, XTIME, 3DTCAM, DRF} have explored accelerating tree-based models using non-volatile CAM, little attention has been paid to the issue of memory redundancy, and existing data placement strategies fail to address this challenge effectively and efficiently. Mitigating redundancy requires a solution that not only reduces model complexity but also optimizes memory utilization, paving the way for a more resource-efficient acceleration.

Based on the above observations, we propose RETENTION, an end-to-end framework that minimizes memory consumption for accelerating tree-based models with CAM. RETENTION offers a pruning algorithm with a novel pruning criterion. The algorithm is applied to bagging-based models iteratively during out-of-bag (OOB) estimation \cite{OOB}, which effectively reduces model complexity while ensuring controlled accuracy degradation. In addition, after analyzing the tradeoffs between different data placement strategies, RETENTION incorporates a tree mapping scheme with two innovative data placement strategies. The strategies are tailored for different optimization criteria, aiming to alleviate memory redundancy and further reduce CAM capacity requirement. 

To the best of our knowledge, this work is the first to explicitly consider the tradeoffs between memory redundancy and processing overhead,  offering data placement strategies for different optimization criteria. RETENTION is evaluated on Random Forest and XGBoost using five datasets. Although validated with these two models, the framework can be generalized to other ensemble models with similar structures such as LightGBM \cite{lightGBM} and CatBoost \cite{CatBoost}. Experimental results show that our tree mapping scheme alone reduces CAM capacity requirement by $1.46\times$ to $21.30\times$, while the full RETENTION framework with model pruning and mapping optimization achieves $4.35\times$ to $207.12\times$ reduction with less than 3\% accuracy degradation. These results demonstrate that RETENTION effectively reduces CAM capacity requirement, making tree-based model acceleration with CAM more resource-efficient and feasible for resource-constrained environments. 

The rest of this paper is organized as follows: Section~\ref{sec:background} presents the background, observation, and motivation of this work. Section~\ref{sec:RETENTION} introduces the philosophy and detailed design of RETENTION. Section~\ref{sec:evaluation} evaluates RETENTION’s effectiveness and compares it with existing works. Finally, Section~\ref{sec:conclusion} provides concluding remarks. 

\section{Background, Observation, and Motivation} \label{sec:background}

\subsection{Decision Tree, Bagging, and Boosting} \label{bg:trees}
\jb{a simple opinion, I think this background part is a way too long (the one related to the trees, for the TCAM part it is ok}

\begin{figure*}[t]
  \centering
  \includegraphics[width=\linewidth, page=1]{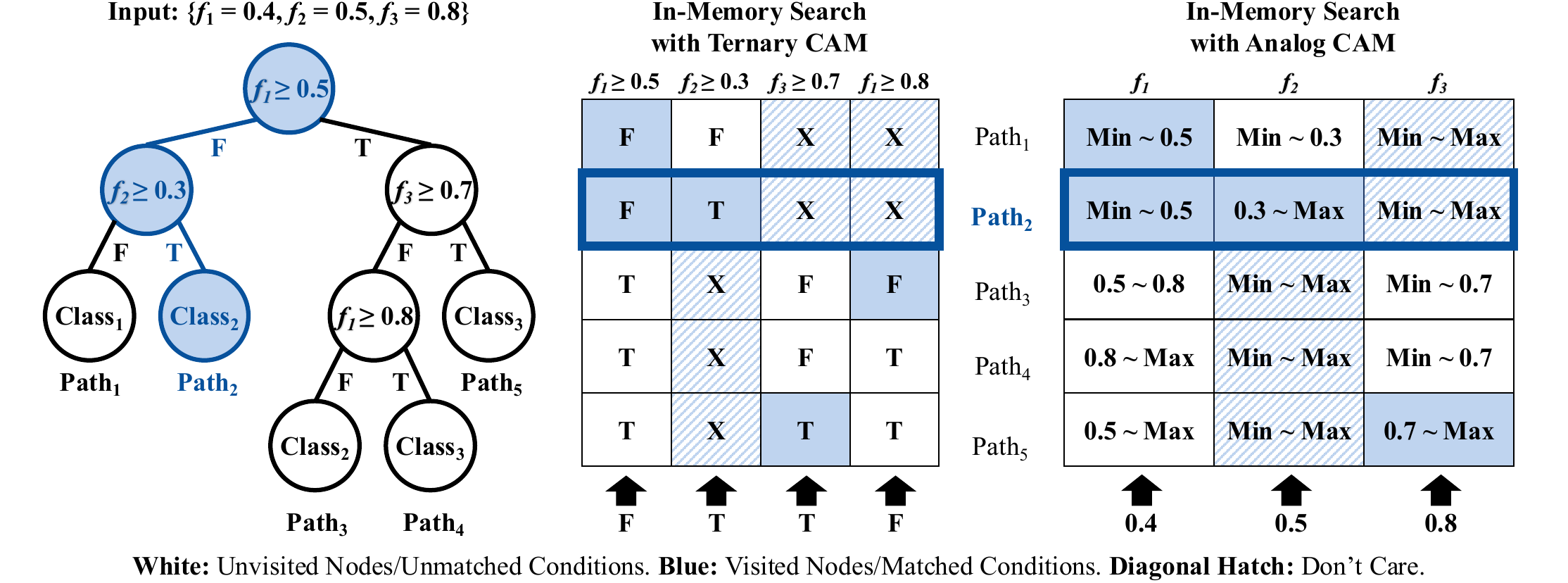}
  \caption{Visualization of decision tree inference acceleration with TCAM and ACAM.}
  \label{fig:CAM}
\end{figure*}

Decision tree \cite{DT} is a commonly used supervised machine learning algorithm for classification and regression tasks, valued for its simplicity in training and high interpretability. Typically structured as a binary tree, each internal node represents a decision condition based on a feature and a corresponding threshold, while each leaf node stores a predicted result. The training process involves a series of node-splitting operations, where all instances initially start at the root node. During each split, instances within the current node are evaluated, and the optimal feature-threshold pair that best separates them (i.e., generates maximum impurity decrease) is selected as the node’s condition. The instances are then distributed to the appropriate child nodes based on whether they satisfy the condition. This process continues iteratively until all instances are separated or predefined constraints, such as maximum depth or minimum impurity decrease, are reached. The left part of Fig.~\ref{fig:CAM} presents an example of decision tree inference. As shown in Fig.~\ref{fig:CAM}, inference follows a similar tree traversal process, where an instance starts at the root node and follows a path determined by the conditions at each encountered node. The prediction result is the value stored in the reached leaf node. Since each root-to-leaf path is determined by a set of conditions that an instance must satisfy during traversal, it can be viewed as a sequence of condition checks leading to the final prediction. Additionally, as there is no contradiction between the conditions within a path, the order of condition checks is irrelevant to the prediction result, providing an opportunity to accelerate inference with CAM.

However, the simplicity in training comes with a major drawback: overfitting, where the model captures noise in the training data, reducing its ability to generalize to unseen instances. To mitigate overfitting and enhance model performance, decision-tree-based ensemble models were introduced. 

Bagging (i.e., bootstrap aggregation) \cite{Bagging} is a widely used algorithm for tree-based ensemble model training. It trains multiple fully-grown decision trees independently, each on a unique subset of the training set generated through bootstrapping (i.e., sampling with replacement). During inference, each tree produces an individual result, and the ensemble determines the final output via majority voting (if classification) or averaging (if regression). By training trees on different subset, bagging enhances robustness against overfitting, reducing the need for pruning. When additional model compression is required, early-stopping (i.e., pre-pruning) is preferred to pruning constructed trees (i.e., post-pruning), as conventional post-pruning algorithms ignore the collective behavior of the model and can erode the diversity that bagging-based model relies on. Additionally, bagging enables OOB estimation, an inherent validation method that eliminates the need for a separate validation set. Since each tree is trained on a subset of data, the unused instances can serve as the validation set, which is later passed through the corresponding tree for evaluation. By aggregating predictions from all trees that did not train on a given instance, OOB estimation provides an unbiased measure of model performance, making it particularly advantageous in data-constrained scenarios. Bagging alleviates overfitting, while the ensemble compensates for the missing information in individual trees, resulting in improved accuracy compared to a single decision tree. Additionally, since each tree in a bagging-based model is trained independently and contributes equally to the final prediction, models such as Random Forest perform exceptionally well on noisy and small datasets. However, they struggle to capture complex patterns, as they do not fully exploit interactions among trees within the ensemble. Moreover, the averaging mechanism limits model performance in regression tasks, highlighting the need for an alternative.

Another class of tree-based ensemble model primitives is the boosting-based model, which is commonly utilized for both classification and regression tasks. In contrast to bagging, which trains multiple fully-grown decision trees independently, boosting \cite{Boosting} trains shallow decision trees sequentially, with each tree attempting to correct the errors of its predecessors. During training, boosting fits the first tree on the entire dataset, and each subsequent tree is trained to correct the errors of the previous ones, with errors being reevaluated at each step as new trees are added to the ensemble. When making a prediction, the final result is computed as a weighted sum of individual tree predictions rather than relying on majority voting or averaging, as trees in boosting-based models contribute with different importance weights. While the error correction mechanism and the sequential training process adapt the models to task complexity, they are more susceptible to overfitting and therefore rely heavily on model pruning. However, since the training phase leverages interactions among trees, post-pruning can disrupt these inter-tree dependencies, leading to catastrophic accuracy degradation. Consequently, boosting-based models often opt for pre-pruning algorithms such as limiting maximum depth and minimum impurity decrease. Nevertheless, modern boosting implementations, such as XGBoost, can easily contain thousands of trees, resulting in substantial memory consumption. Additionally, the irregular tree structures and non-deterministic memory access patterns make efficient inference on conventional hardware particularly challenging. These limitations underscore the potential of CAM, which replaces the time- and energy-consuming memory accesses with efficient in-memory search.

\subsection{Ternary CAM and Analog CAM} \label{bg:CAM}

Content-addressable memory (CAM) \cite{SRAM, CAM, CAMSurvey} is a specialized memory architecture widely used in various applications. Unlike conventional memory, which retrieves data based on a given address, CAM allows simultaneous comparison of an input query sequence against all stored sequences, and returns the addresses or associated data of matching entries. Data in CAM is stored row-wise, with comparisons performed in parallel on a column basis. This high degree of parallelism makes CAM particularly well-suited for applications requiring rapid lookups, such as network routing, database indexing, and cache systems. \jb{do you think a figure of the CAM architecture would be useful ? also I think it is good to clearly discuss the maturity of CAMs in one or 2 sentences}

Ternary content-addressable memory (TCAM) \cite{TCAM} extends the functionality of binary CAM by introducing a third state, \textit{don't care}, denoted as X. Bits set to the X state are treated as matches during comparison, enabling greater flexibility in search operations. Since each root-to-leaf path in a tree-based model represents a sequence of conditions an input must satisfy to yield the corresponding prediction, TCAM can efficiently traverse these paths, making it a compelling solution for tree-based model acceleration. Prior work \cite{DT2CAM} proposed mapping each path to a TCAM row, with each column representing a unique condition. Unencountered conditions of a path are assigned the X state as they do not affect the traversal process. When an input instance is received, it is first encoded into a binary sequence representing the condition check results. This  sequence is then used to construct TCAM input queries, which are dispatched to the respective TCAMs for in-memory search. The middle part of Fig.~\ref{fig:CAM} illustrates how TCAM serves as a decision tree accelerator. The decision tree at the left part of Fig.~\ref{fig:CAM} is mapped to the TCAM, where four unique conditions are assigned to separate columns, and each path is represented as a row. Conditions are encoded as 0, 1, and X, corresponding to \textit{False}, \textit{True}, and \textit{don't care}, respectively. Once an input is received, the features are encoded into a binary sequence representing the condition check results, and each bit is then delivered to the corresponding column to perform in-memory search. Since all cells in the second row are matched, $Path_2$ is the traversal path for the input instance.


\begin{table*}[t]          
\centering                 
\caption{dataset information and the experiment results for significant redundancy.}
\label{tab:redundancy}
\renewcommand{\arraystretch}{1.1}
\begin{tabular}{c|cccc|cccccc}    
\toprule                    
Dataset & \#Samples & \#Features & \#Classes & Attribute & \#Paths & Avg. Length & \#Unique Cond. & Size (MB) & Redundancy (\%)\\
\midrule
Adult \cite{Adult} & 48842 & 14 & 2 & medium & 206152 & 16.91 & 29436 & \textbf{723.40} & \textbf{99.94}\\
CreditApproval \cite{CreditApproval} & 690 & 15 & 2 & small &   6246 &  7.39 &  1934 &   1.44 & \textbf{99.61}\\
DryBean \cite{DryBean} & 13611 & 16 & 7 & multi-class &  52663 & 12.11 & 41752 & 262.12 & \textbf{99.97}\\
Letter \cite{Letter} & 20000 & 16 & 26 & multi-class & 190377 & 15.59 &     421 &   9.55 & \textbf{96.30}\\
Wine \cite{Wine} & 4898 & 11 & 11 & multi-class & 107772 & 13.53 &   5467 &  70.24 & \textbf{99.75}\\
\bottomrule
\end{tabular}
\end{table*}

Analog content-addressable memory (ACAM) \cite{ACAM} follows a similar structural design to TCAM, but differs in its data representation. Instead of discrete states (0, 1, X), each ACAM cell can store a range of analog values. A cell is considered matched if the given input value falls within the stored range, meaning that a min-to-max range in ACAM functions analogously to the X state in TCAM. Previous research \cite{TreeBasedACAM} has demonstrated the potential of ACAM for accelerating Random Forest. Similar to \cite{DT2CAM}, each root-to-leaf path is mapped to an ACAM row. However, unlike TCAM, where each column represents a feature-threshold condition, ACAM columns directly correspond to features. When an input instance is received, a digital-to-analog converter first transforms the features into analog voltages. ACAM then performs in-memory search operations to retrieve the final result. The right part of Fig.~\ref{fig:CAM} provides an overview of ACAM serving as an accelerator for the decision tree in the left part of Fig.~\ref{fig:CAM}. In contrast to TCAM, ACAM only requires three columns to represent the features, and each of the cells in ACAM stores an analog range instead.

However, despite its lower capacity requirement, ACAM demands additional peripheral circuits, and its larger cell size does not necessarily offer better area efficiency than TCAM. Furthermore, analog computing is susceptible to various non-idealities, such as stuck-at faults, IR drop, thermal noise, shot noise, and random telegraph noise \cite{NonIdealities}. While mitigation techniques exist \cite{Robustness}, they often introduce additional memory overhead and offer limited effectiveness. Due to these reliability challenges, ACAM is not yet viable for real-world implementation. Therefore, this work opts for TCAM as the underlying hardware. Still, RETENTION remains compatible with ACAM after slight modifications (refer to Section~\ref{method:generalizability} for detailed explanation).

\subsection{Observation}

To accelerate tree-based model inference using TCAM, each root-to-leaf path is mapped to a TCAM row, with each column representing a unique condition within the ensemble. Since a single path encounters only a minimal fraction of the conditions within the ensemble, the majority of TCAM cells store the X state for format alignment, leading to significant redundancy. In fact, mapping a tree-based model to TCAM without optimizations requires at least $ \#\textit{paths} \times \#\textit{unique\_conditions} $ bits, which might exceed the memory capacity available in resource-constrained environments. As shown in Fig.~\ref{fig:CAM}, mapping a decision tree to a CAM with perfect capacity leads to over one-third of cells storing the X state. Moreover, due to the extremely high write latency of nvTCAM, real-time writing severely degrades performance, making it critical to fit the entire model into nvTCAM. Table~\ref{tab:redundancy} presents an experiment that highlights the substantial memory requirement and the overwhelming redundancy caused by the widespread use of the X state in TCAM. In this experiment, Random Forest models with 100 decision trees require up to 700MB of memory when mapped with \textit{naive unified mapping} (refer to Section~\ref{method:naive} for detailed explanation), and all exhibited redundancy of more than 96\%. These results highlight that directly deploying nvTCAM as a tree-based model accelerator is both infeasible and inefficient in resource-constrained environments.

Previous work DT2CAM \cite{DT2CAM} introduced a framework for mapping decision trees to TCAM. However, instead of addressing memory redundancy, it only proposed an energy-saving precharge mechanism that disables mismatched rows to conserve power. Pedretti et al. \cite{TreeBasedACAM} proposed mapping Random Forests to ACAM, incorporating techniques such as (1) limiting maximum tree depth and (2) reordering features while eliminating empty rows to mitigate memory redundancy. However, pre-pruning algorithms can result in severe accuracy degradation when implementing on bagging-based models (refer to Section~\ref{method:pruning} for detailed explanation), and alternative pruning strategies could achieve better accuracy within the same memory budget. Furthermore, although feature reordering groups redundant cells and empty-row elimination reduces capacity requirement, the associated computational overhead for merging partial match results is relatively high. In fact, comparable computational overhead could yield even greater memory savings. A follow-up study by Pedretti et al. \cite{XTIME} proposed an ACAM-based analog-digital architecture for tree-based ensemble model acceleration, assigning individual trees to separate ACAMs for architectural and compilation simplicity. While this approach effectively reduces redundancy across different trees, redundancy within each separate tree remains, leaving room for further optimization. Addressing this residual redundancy could lead to more reduction in memory consumption while maintaining computational efficiency.

\subsection{Motivation}

This work is driven by the challenges identified from the previous works mentioned above. While nvTCAM holds great promise for accelerating tree-based model inference in resource-constrained environments, the excessive CAM capacity requirement remains a critical challenge that must be addressed before practical implementation. Therefore, this paper proposes RETENTION, an end-to-end framework that significantly reduces CAM capacity requirement, enabling resource-efficient tree-based ensemble model acceleration.

\section{RETENTION} \label{sec:RETENTION}

\begin{figure*}[t]
  \centering
  \begin{minipage}[t]{0.49\textwidth}
    \centering
    \includegraphics[width=\textwidth, page=2]{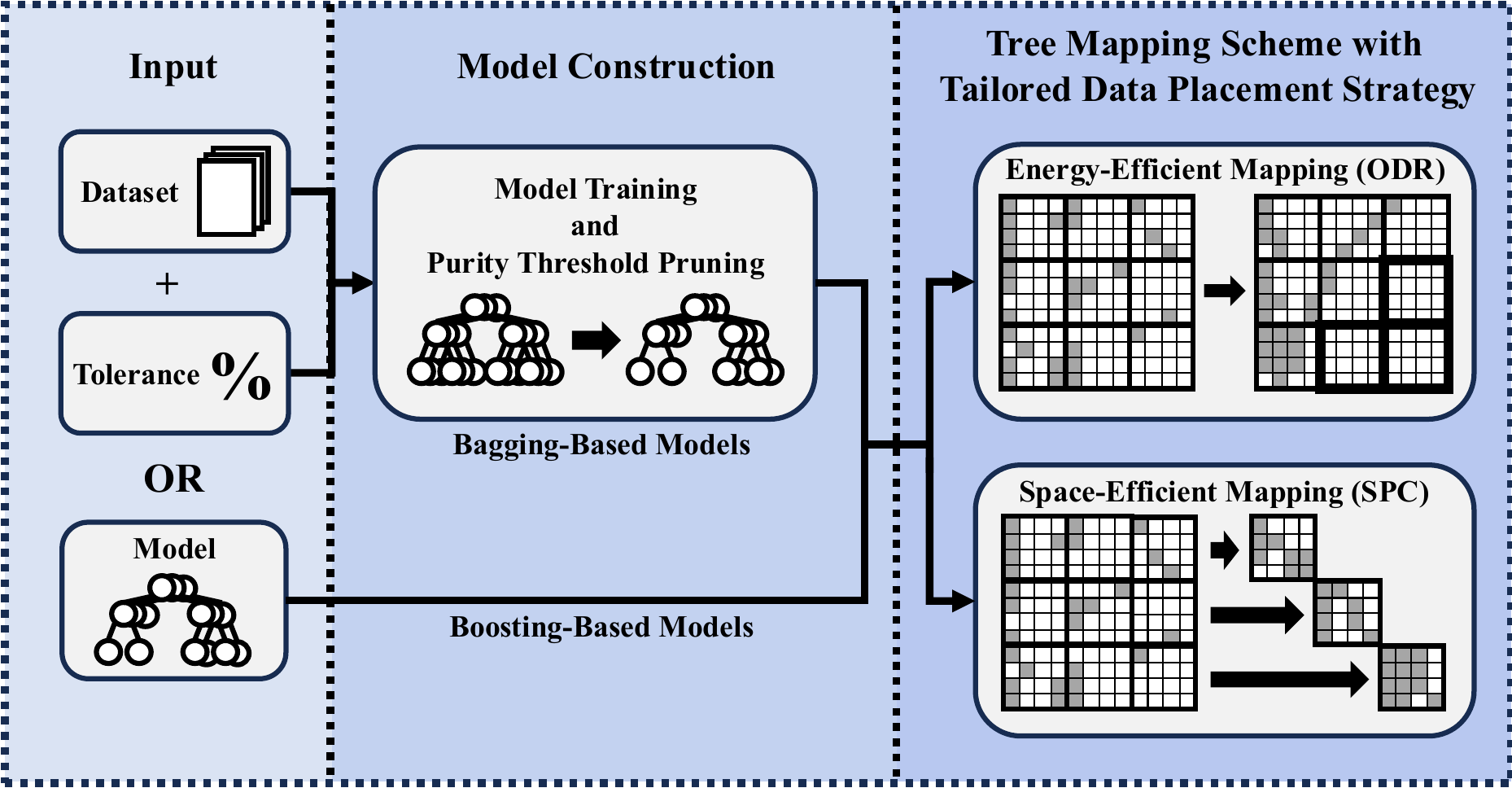}
    \caption{Overview of RETENTION.}
    \label{fig:overview}
  \end{minipage}
  \hfill
  \begin{minipage}[t]{0.49\textwidth}
    \centering
    \includegraphics[width=\textwidth, page=3]{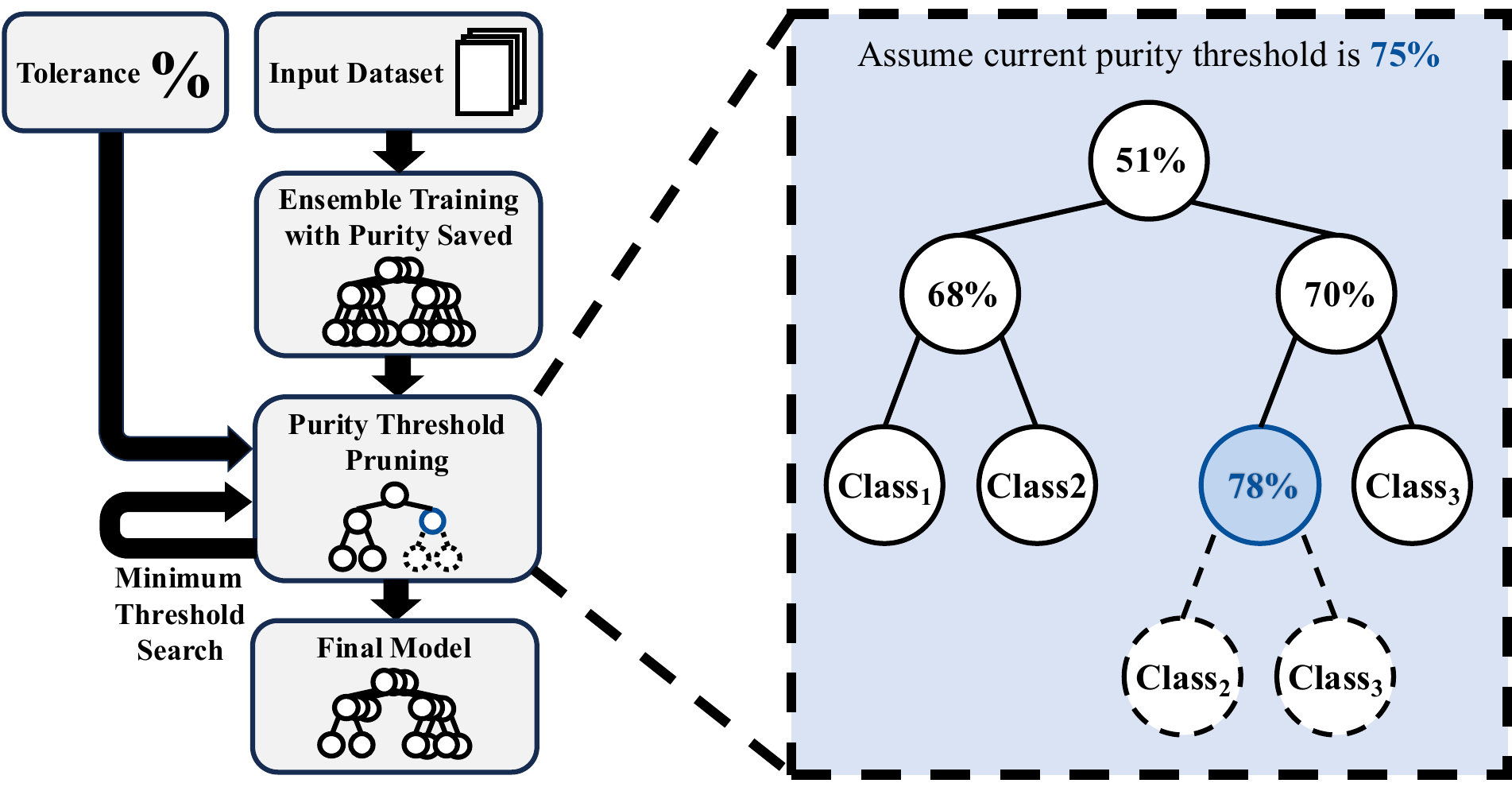}
    \caption{Workflow of model construction with purity threshold pruning.}
    \label{fig:pruning}
  \end{minipage}
\end{figure*}

\subsection{Overview}

nvTCAM enables high-speed, energy-efficient inference for tree-based models, making it a promising solution for acceleration. However, substantial memory consumption and redundancy remain key barriers to real-world deployment, limiting its feasibility in resource-constrained environments. In this section, we introduce RETENTION, an end-to-end framework serving as a crucial component to enable practical and resource-efficient acceleration of tree-based model inference with nvTCAM. It addresses the challenges by (1) minimizing model complexity through \textit{purity threshold pruning} and (2) enhancing memory utilization by introducing an optimized tree mapping scheme with two data placement strategies, namely \textit{occurrence-based double reordering (ODR)} and \textit{similarity-based path clustering (SPC)}. Fig.~\ref{fig:overview} presents an overview of RETENTION, detailing its key components and their interactions. The following sections elaborate on the philosophy and implementation of each component, concluding with a discussion on the generalizability of RETENTION to ACAM.

\subsection{Purity Threshold Pruning} \label{method:pruning}

As the required CAM capacity is correlated to both $\#\textit{paths}$ and $\#\textit{unique\_conditions}$, pruning models can achieve effective mitigation by reducing both at the same time. Existing pruning techniques for reducing tree-based ensemble model complexity mainly focus on pre-pruning (as explained in Section~\ref{bg:trees}), and the algorithms fall into two main categories: (1) structural restriction, which limits the tree’s shape with hard constraints (e.g., maximum depth), and (2) split-based pruning, which stops splitting when subsequent splits are deemed worthless (e.g., minimum impurity decrease). While both methods reduce model complexity, neither takes into account the class distribution of instances within nodes, potentially leading to significant accuracy degradation. For example, in complex datasets, instances from different classes may remain entangled even at the maximum depth, degrading accuracy substantially when forced truncation occurs. Similarly, if an early split does not effectively separate instances, later splits might still provide meaningful differentiation, yet split-based pruning prematurely terminates such opportunities. While these techniques have minimal impact on boosting-based models, where trees iteratively correct previous errors, bagging-based models can suffer severe accuracy loss due to their independently trained trees, which lack an error correction mechanism.

To avoid potential catastrophic accuracy degradation, this work introduces \textit{purity threshold pruning}, a novel pruning algorithm designed for bagging-based models. The workflow is depicted in the left part of Fig.~\ref{fig:pruning}. Unlike conventional training, which only takes a dataset as input, our approach incorporates a user-specified tolerance for OOB accuracy loss. During training, each node records its majority class and the corresponding purity (i.e., proportion). Once the ensemble is constructed, rather than concluding with OOB estimation, we iteratively determine the minimum purity threshold that maintains accuracy within the specified tolerance. Nodes exceeding this threshold are converted into leaf nodes representing their respective majority class, as shown in the right part of Fig.~\ref{fig:pruning}.

\begin{table*}[t]
    \centering
    \renewcommand{\arraystretch}{1.35}
    \caption{comparison of naive unified mapping and naive independent mapping.}
    \label{tab:mapping_comparison}
    \begin{tabular}{c|c|c}
        \toprule
        \textbf{Attribute} & \textbf{Naive Unified Mapping} & \textbf{Naive Independent Mapping} \\
        \midrule
        \textbf{Mapping Unit} & entire ensemble & single tree \\
        
        
        
        \textbf{Total \#TCAM Required} & 
        $\left\lceil \frac{\#\textit{unique\_conditions}}{S} \right\rceil \times \left\lceil \frac{\#\textit{paths}}{S} \right\rceil$ & 
        $\sum \left( \left\lceil \frac{\#\textit{unique\_conditions}}{S} \right\rceil \times \left\lceil \frac{\#\textit{paths}}{S} \right\rceil \right)$ \\
        
        \textbf{Key Strength} & lower input preprocessing cost & lower capacity requirement \\
        \textbf{Category} & \textbf{energy-efficient mapping} & \textbf{space-efficient mapping} \\
        \bottomrule
    \end{tabular}
\end{table*}

The core principle of \textit{purity threshold pruning} is to reduce model complexity while preserving accuracy at the ensemble level. Since individual trees are trained on different subsets of training data with distinct splits, a minority class overlooked in one tree can still be classified correctly in others. For highly imbalanced datasets, applying class weighting helps prevent the model from favoring majority classes. Additionally, by explicitly considering OOB accuracy during pruning, the method ensures that performance degradation remains controlled. Unlike standard pruning techniques that may create leaf nodes with evenly distributed classes, potentially leading to misclassifications, \textit{purity threshold pruning} ensures that each leaf node maintains sufficient class purity to guarantee the desired level of accuracy. Moreover, when early splits already satisfy the target accuracy, further splits are omitted, significantly reducing both $\#\textit{paths}$ and $\#\textit{unique\_conditions}$ compared to conventional pruning methods. Different from most post-pruning algorithms that operate on individual decision trees in isolation, \textit{purity threshold pruning} is specifically designed to account for the collective behavior of ensemble models during inference. This global perspective enables \textit{purity threshold pruning} to preserve diversity and effectively reduce model complexity while explicitly constraining accuracy degradation within a user-defined tolerance, a capability not possessed by conventional single-tree pruning methods.

Recent innovations in pruning focus on removing entire trees from models if they show substantial redundancy or structural resemblance \cite{PruneWholeTree1, PruneWholeTree2}. Since \textit{purity threshold pruning} operates at the node level and assumes trees are diversely grown, it is fully compatible with such whole-tree pruning methods, exhibiting high adaptability for resource-constrained environments. However, it is not compatible with boosting-based models, as pruning such models after construction ruins their error correction mechanism, leading to catastrophic accuracy degradation. Still, the sequential training of boosting-based models allow existing pre-pruning algorithms to sufficiently reduce complexity without compromising accuracy.

\subsection{Input Pre-Processing during Inference} \label{method:preprocessing}

To accelerate tree-based model inference with CAM, raw input data must be transformed into the specific query format for individual search operations. This transformation involves two key steps: (1) feature encoding and (2) query packing. For feature encoding, the thresholds of each feature must be sorted in advance for efficient inference. When raw input data for a feature is received, binary search can be employed to accelerate threshold comparisons. This step can be performed on the sensor side in the context of near-sensor inference, as modern smart sensors are capable of handling such lightweight tasks \cite{SmartSensor}. Additionally, since each feature is typically associated with a dedicated sensor, pre-processing at the sensor level is feasible and efficient. Once features are encoded into binary sequences, the query packing step organizes condition check results into the query format, ensuring alignment with the order of TCAM columns. These queries are then transmitted to the corresponding TCAM via a network-on-chip (NoC) \cite{NoC}, following the same implementation as in \cite{XTIME}.

Since tree traversal is essentially a sequence of condition checks, encoding input features into sequences and performing in-memory search may initially seem redundant. However, due to the frequent condition sharing across the ensemble, the actual number of unique conditions is significantly lower than expected. Moreover, binary search can further reduce the number of condition checks for feature encoding. For instance, in a Random Forest model with 100 trees, 14 input features, and an average path length of 17.38, traversing the forest with conventional processors requires about 1738 condition checks. On the other hand, experimental result reveals that there are 5446 unique conditions within the model. Applying binary search can reduce the number of condition checks to approximately 120 ($14~\textit{features} \times \log_2\frac{5446}{14}$), which is a significant improvement. Additionally, feature encoding is highly efficient compared to conventional CPU-based tree traversal, as all data accesses are deterministic and easily cached. Given that the computational overhead for feature encoding is constant when using TCAM as an accelerator, this work focuses on the computational overhead of query packing, which is determined by the selected data placement strategy. \jb{same as previous comment, maybe we  can have a summary through an algorithm. If I was reviewer, I would have appreciated a more synthetic description of the steps performed here}

\begin{figure*}[t]
  \centering
  \includegraphics[width=\linewidth, page=4]{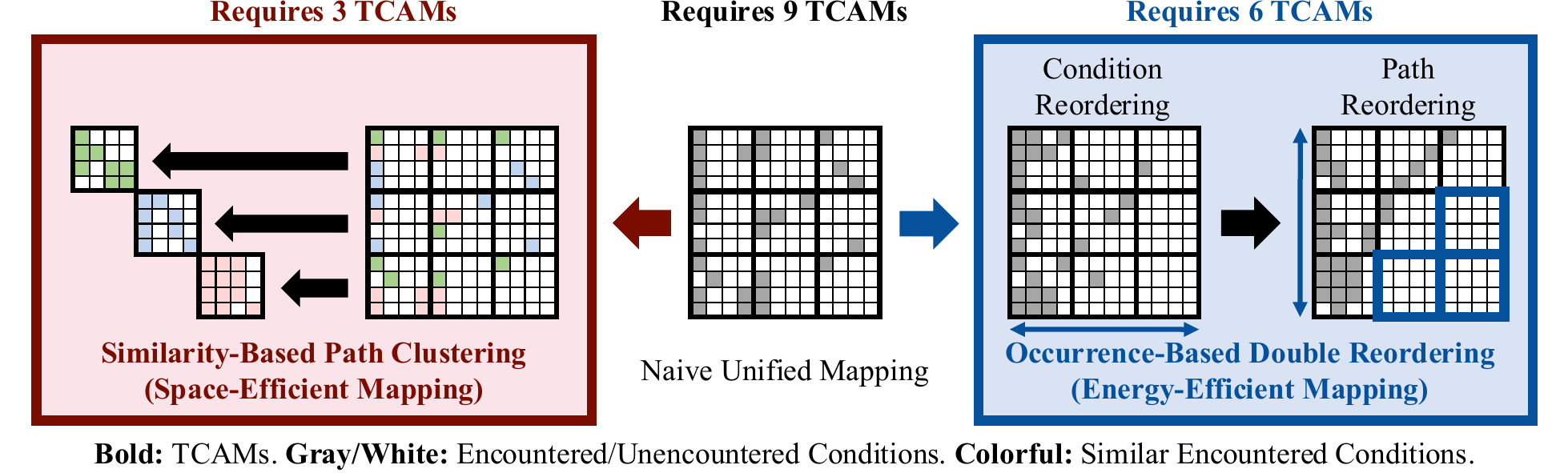}
  \caption{Visualization of the proposed data placement strategies.}
  \label{fig:placement}
\end{figure*}

\subsection{Tree Mapping Scheme} \label{method:mapping}

\subsubsection{Naive Tree Mapping Approaches} \label{method:naive}

To utilize TCAM for acceleration, each root-to-leaf path within the model needs to be mapped to a TCAM row, with every column of the TCAM corresponding to a unique condition within the paths being searched. There are two naive approaches for mapping: (1) \textit{naive unified mapping} and (2) \textit{naive independent mapping}, with their differences summarized in Table~\ref{tab:mapping_comparison}, assuming a TCAM size of $\textit{S} \times \textit{S}$. In \textit{naive unified mapping}, paths from the entire ensemble are mapped together, so that each row entry corresponds to the same set of conditions. This structure allows for efficient input pre-processing since TCAMs in the same column receive identical input queries. However, most nodes in an ensemble are only encountered by few paths, resulting in a vast number of cells storing the X state. In fact, mapping 100 trees together can lead to over 96\% redundancy, as shown in Table~\ref{tab:redundancy}. Conversely, \textit{naive independent mapping} maps each tree separately, ensuring that conditions unique to one tree do not introduce redundancy to rows associating with paths in other trees. Although this approach significantly alleviates redundancy, each TCAM requires a distinct input query, increasing pre-processing overhead.

While \textit{naive unified mapping} generally demands more TCAMs than \textit{naive independent mapping}, the energy-saving precharge mechanism proposed in DT2CAM \cite{DT2CAM} can help mitigate energy waste by deactivating mismatched rows early. Furthermore, given that input pre-processing dominates energy consumption, as CPU computation is considerably more power-intensive than in-memory search operation (refer to Table~\ref{tab:energy} for detailed statistics), and that  \textit{naive unified mapping} requires fewer iterations of query packing due to query sharing, it is more energy-efficient than \textit{naive independent mapping}. Therefore, based on the above observations, we categorize the optimization criteria of data placement strategies into two types: energy-efficient mapping and space-efficient mapping. The key difference is whether to minimize capacity requirement at the cost of increased computational overhead.

\subsubsection{Existing Data Placement Strategy Analysis} \label{method:existing}

DT2CAM primarily focused on mapping a single decision tree to TCAM without addressing redundancy or ensemble models. When directly applied to ensembles, DT2CAM acts as either \textit{naive unified mapping} or \textit{naive independent mapping}, depending on the chosen mapping unit. Pedretti et al. \cite{TreeBasedACAM} introduced a data placement strategy, referred to as the \textit{feature reordering with row elimination (FR)} algorithm, for ACAM. This method selects the whole ensemble as a mapping unit, and proposes (1) reordering features based on frequency to cluster redundant cells, and (2) eliminating entire path segments in each ACAM when they are fully redundant. However, due to row elimination, paths were no longer stored in fixed rows across different ACAMs, making the merging of partial match results substantially more complex and CPU-dependent. As a result, the additional computational overhead is proportional to $\#\textit{ACAMs}$, similar to that of \textit{naive independent mapping} for query packing. Thus, we classify \textit{FR} as a space-efficient mapping approach, though its capacity requirement may be higher than \textit{naive independent mapping}. A follow-up study by Pedretti et al. \cite{XTIME} proposed fitting each tree into separate ACAMs, which is identical to \textit{naive independent mapping}.

After a thorough analysis of the tradeoffs among existing data placement strategies, we find that prior approaches fail to effectively alleviate memory redundancy, offering efficiency comparable to naive methods. To bridge this gap, we introduce two novel data placement strategies: one optimized for energy-efficient mapping and the other for space-efficient mapping. The following sections (~\ref{subsec:ODR}) provide a detailed explanation of these strategies, with a visualization presented in Fig.~\ref{fig:placement}.

\subsubsection{Occurrence-Based Double Reordering (ODR)} \label{subsec:ODR}

To alleviate memory redundancy in an energy-efficient manner, each path should be stored in a fixed row to avoid additional computational overhead when merging partial match results. Moreover, input queries for TCAMs in the same column must remain consistent to minimize input pre-processing overhead for query packing. Given these constraints, we propose \textit{ODR} to enhance memory utilization for resource-constrained environments that prioritize energy efficiency. As outlined in Algorithm~\ref{alg:odr}, \textit{ODR} first sorts conditions by occurrence frequency in descending order, similar to \textit{FR}. Instead of eliminating partial sequences when a TCAM row is entirely filled with X states, which introduces sequential computational overhead as seen in \textit{FR}, \textit{ODR} optimizes path ordering based on condition usage. Paths that contain rare conditions are placed at the top, concentrating most of the cells storing the X state in the bottom-right TCAMs. TCAMs that are utterly filled with X states can then be removed, reducing redundancy without additional computational overhead.

Despite its effectiveness, \textit{ODR} still leaves considerable redundancy, as shown in Fig.~\ref{fig:placement}. For example, there is only one meanful cell in the top-right TCAM after applying \textit{ODR}, yet the entire TCAM is required for correct functionality. To further reduce capacity requirement, the next section introduces a more aggressive data placement strategy that reduces redundancy at the cost of slightly higher energy consumption. However, its computational overhead is comparable to that of \textit{naive unified mapping}, and is, in fact, even lower in practice.

\begin{algorithm}[t]
    \caption{Occurrence-Based Double Reordering (ODR)}
    \label{alg:odr}
    \begin{algorithmic}[1] 
        \Require \textit{pool} = root\_to\_leaf\_paths
        \Require \textit{conditions} = pool.union()
        \Ensure \textit{condition\_order}, \textit{path\_order}
        
        \State \textit{paths} $\gets [\ ]$ 
        \State \textit{sort(conditions, descending)}
        
        \For{\textit{c} in \textit{conditions}.reverse()} 
            \For{\textit{path} in \textit{pool}}
                \If{$c \in path$}
                    \State \textit{paths}.append(\textit{path})
                    \State \textit{pool}.remove(\textit{path})
                \EndIf
            \EndFor
        \EndFor
        
        \State \Return \textit{conditions}, \textit{paths}
    \end{algorithmic}
\end{algorithm}

\subsubsection{Similarity-Based Path Clustering (SPC)}

Since memory redundancy arises from unencountered conditions, clustering paths with high similarity (i.e., those that share many conditions) can substantially reduce capacity requirement. Although \textit{naive independent mapping} can be viewed as a basic form of clustering, paths within a tree exhibit limited similarity, and some TCAMs can still remain nearly redundant especially in cases where $\#\textit{paths} = \textit{S} + 1$ and the TCAM size is $\textit{S} \times \textit{S}$, leading to inefficient capacity utilization. Finding the optimal clusters to minimize TCAM consumption is an NP-hard problem. In this section, we propose \textit{SPC}, a heuristic algorithm that greedily maximizes similarity among paths within each TCAM, and thus minimizes capacity requirement. As presented in Algorithm~\ref{alg:spc}, \textit{SPC} starts with a \textit{pool} containing all root-to-leaf paths in the model. While \textit{pool} is not empty, \textit{SPC} evaluates the similarity between \textit{current\_cluster} and each of the paths within \textit{pool}, and designates the path that can fit in \textit{current\_cluster} based on (1) the highest number of shared conditions, and (2) the smallest resulting set of unique conditions after being integrated into \textit{current\_cluster}, as \textit{candidate}. Since the objective is to accommodate all paths using the smallest number of clusters, and the capacity of each cluster depends on both $\#\textit{paths}$ and $\#\textit{unique\_conditions}$ because we limit a cluster to a single TCAM, the key intuition behind \textit{SPC} is to minimize $\#\textit{unique\_conditions}$ at every step. This approach allows each TCAM to hold as many paths as possible, resulting in an effective heuristic. Once \textit{current\_cluster} reaches capacity or none of the paths is suitable, \textit{current\_cluster} is added to \textit{clusters} representing the completed ones, and a new empty cluster is initiated. Otherwise, \textit{SPC} greedily adds the candidate to \textit{current\_cluster} and removes it from \textit{pool}, and this process is repeated until all paths have been assigned to clusters.

\textit{SPC} efficiently exploits similarity across paths, diminishing unnecessary X states. By filling TCAMs as fully as possible, it maximizes utilization and minimizes wasted capacity. Since each cluster is constrained to contain at most \textit{S} unique conditions and \textit{S} paths, there is no dependency between the match results of different TCAMs, eliminating any computational overhead associated with merging partial match results. Therefore, the only computational overhead lies in the query packing for every TCAM, which is the same as \textit{naive independent mapping}. However, since \textit{SPC} requires far fewer TCAMs than \textit{naive independent mapping}, it emerges as a significantly more resource-efficient approach in terms of both energy and memory consumption.

\begin{algorithm}[t]
    \caption{Similarity-Based Path Clustering (SPC)}
    \label{alg:spc}
    \begin{algorithmic}[1] 
        \Require \textit{S} = TCAM\_size 
        \Require \textit{pool} = root\_to\_leaf\_paths 
        \Ensure \textit{clusters}
        
        \State \textit{clusters} $\gets [\ ]$, \textit{curr\_cluster} $\gets [\ ]$
        
        \While{\textit{pool} is not empty}
            \State \textit{similarity} $\gets$
            calc\_similarity(\textit{pool}, \textit{curr\_cluster}) 
            \State \textit{candidate} $\gets$ find\_best\_candidate(\textit{similarity}, \textit{pool}) 
            
            \If{\textit{candidate} is None \textbf{or} \textit{curr\_cluster.size()} == \textit{S}}
                \State \textit{clusters}.append(\textit{curr\_cluster}) 
                \State \textit{curr\_cluster} $\gets \{\}$ 
            \Else
                \State \textit{curr\_cluster}.append(\textit{candidate}) 
                \State \textit{pool}.remove(\textit{candidate})
            \EndIf
        \EndWhile
        \If{\textit{curr\_cluster} is not empty}
                \State \textit{clusters}.append(\textit{curr\_cluster}) 
        \EndIf

        \State \Return \textit{clusters}
    \end{algorithmic}
\end{algorithm}

\subsection{Generalizability of RETENTION to ACAM} \label{method:generalizability}

\subsubsection{Input Pre-Processing during Inference}

The input pre-processing for ACAM is functionally equivalent to the TCAM flow, as quantization can also be viewed as a sequence of condition checks. However, since ACAM operates on analog values, digital-to-analog conversion is required once the input query is dispatched to the respective ACAM.

\subsubsection{Tree Mapping Scheme}

The fundamental difference between mapping to TCAM and ACAM lies in the meaning of a column. Since each column represents a feature rather than a condition in ACAM, the condition reordering of \textit{ODR} should be changed to feature reordering, while the similarity of \textit{SPC} should denote the number of shared features. While RETENTION supports ACAM, the relative reduction in capacity requirement is less pronounced when $\#features$ and the model size are both small. This is because ACAM inherently requires fewer columns than TCAM for a given model, leading to diminishing returns for smaller-scale configurations.

\section{Evaluation} \label{sec:evaluation}


\begin{figure*}[t]
  \centering
  \includegraphics[width=\linewidth, page=5]{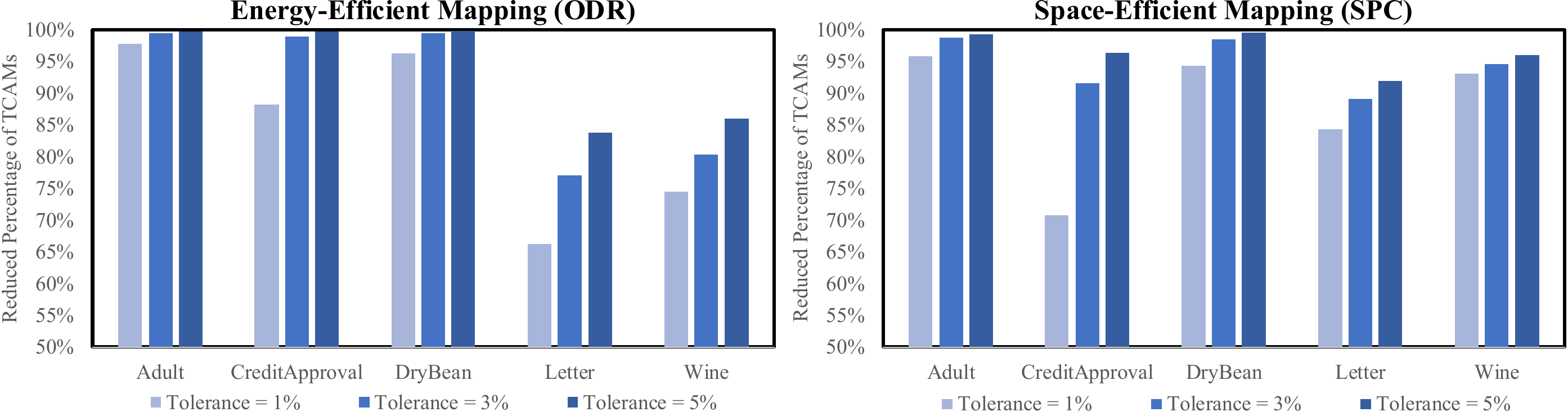}
  \caption{Overall performance of RETENTION.}
  \label{fig:exp_overall}
\end{figure*}

\subsection{Experimental Setup}

\subsubsection{Objective and Metric}
Since previous works \cite{DT2CAM, TreeBasedACAM, XTIME} already demonstrate that CAM outperforms conventional accelerators, in this work we focus on the CAM-oriented optimization, omitting performance comparison with CPU, GPU, FPGA. To evaluate the effectiveness of RETENTION, we compare the number of required TCAMs against previous works, with the reduced number of TCAMs as our evaluation metric. Therefore, higher values represent better performance.

\subsubsection{Model Settings}
Two ensemble model types are selected for case study: Random Forest, representing bagging-based models, and XGBoost, representing boosting-based models. Both types are trained on five datasets from the widely used UCI Machine Learning Repository \cite{UCI}, as listed in Table~\ref{tab:redundancy}. Each dataset is divided into training and testing sets in a 7:3 ratio to examine the impact of \textit{purity threshold pruning} on accuracy. For Random Forest, we utilized the open-source library Ranger \cite{Ranger} for forest construction, setting the number of trees to 100. \textit{Purity threshold pruning} is applied on Random Forest to reduce model complexity, with tolerance set to \{1\%, 2\%, 3\%, 4\%, 5\%\}. On the other hand, we used the official XGBoost library \cite{XGBoost} to construct XGBoost with default parameters, including \textit{max\_depth=6} and \textit{num\_trees=100}. 

\subsubsection{Data Placement Strategies and Baselines}
To achieve resource-efficient acceleration, we implemented \textit{ODR} and \textit{SPC} to further enhance efficiency. As analyzed in Section~\ref{method:existing}, the data placement strategies presented in prior works are mostly equivalent to \textit{naive unified mapping} and \textit{naive independent mapping}, which fall under the categories of energy-efficient mapping and space-efficient mapping, respectively. While \textit{FR} incorporates additional optimizations, the associated computational overhead is comparable to \textit{naive independent mapping}, and is therefore categorized into space-efficient mapping. Since the algorithm is originally designed for ACAM, we make a minor modification, reordering conditions instead of features, to adapt it for TCAM deployment. In this work, energy-efficient mapping and space-efficient mapping are evaluated separately, with \textit{naive unified mapping} and \textit{naive independent mapping} as baselines. 

\subsubsection{Simulation}
The mapping simulation is conducted by storing root-to-leaf paths into tabular format (i.e., each row is a root-to-leaf path, and every column corresponds to a unique condition), and directly calculate the required number of TCAMs, as no open-source simulator currently supports such functionality. All models are mapped to TCAMs with size $64\times64$ if not specified. Both energy consumption and latency are evaluated on Intel Xeon Gold 6530 under a single-threaded configuration, using Linux Perf for energy profiling and the C++ \textit{std::chrono} library for \textit{ns}-resolution timing. To mitigate system noise, all results are averaged over $10^9$ iterations.

The subsequent sections present detailed experimental results: Section~\ref{exp:overall} analyzes RETENTION’s overall performance. Section~\ref{exp:pruning} investigates the effects of \textit{purity threshold pruning} on accuracy and TCAM capacity requirement. Section~\ref{exp:placement} evaluates the contributions of \textit{ODR} and \textit{SPC} individually. Section~\ref{exp:tree} explores the impact of the number of trees. Section~\ref{exp:TCAM} examines the influence of TCAM size. Finally, Section~\ref{exp:energy} discusses RETENTION's influence on energy consumption and latency. Detailed experimental statistics of applying RETENTION are summarized in Table~\ref{tab:RETENTION}.

\subsection{Experiment 1: Overall Performance of RETENTION} \label{exp:overall}

To evaluate the overall performance of RETENTION, we compare Random Forest models with tolerance set to \{1\%, 3\%, 5\%\} against baseline methods. As shown in Fig.~\ref{fig:exp_overall}, RETENTION exhibits substantial reductions, achieving 66.25\% to 99.96\% fewer TCAMs required (equivalent to $2.96\times$ to $2420.81\times$ capacity reduction) in energy-efficient mapping, and 70.83\% to 99.70\% reduction ($3.43\times$ to $334.48\times$ improvement) in space-efficient mapping. Specifically, setting tolerance to a moderate value of 3\% yields $4.35\times$ to $207.12\times$ improvement. These results demonstrate that RETENTION effectively minimizes CAM capacity requirement, offering a promising solution for resource-efficient acceleration with CAM.

\subsection{Experiment 2: Purity Threshold Pruning} \label{exp:pruning}

\begin{figure*}[t]
  \centering
  \includegraphics[width=\linewidth, page=6]{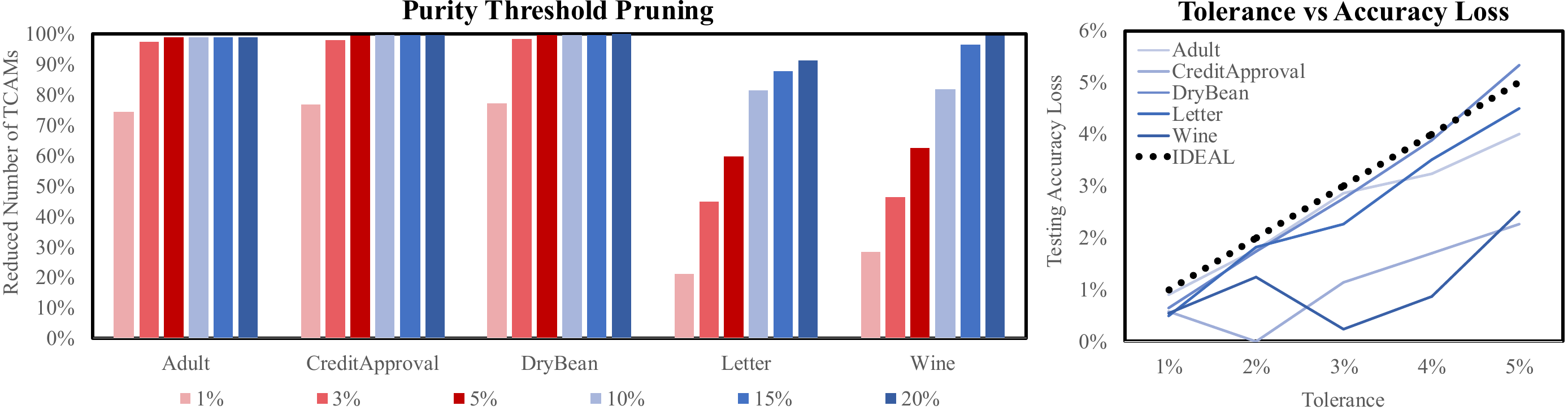}
  \caption{Effectiveness of purity threshold pruning.}
  \label{fig:exp_pruning}
\end{figure*}
\begin{figure*}[t]
  \centering
  \includegraphics[width=\linewidth, page=7]{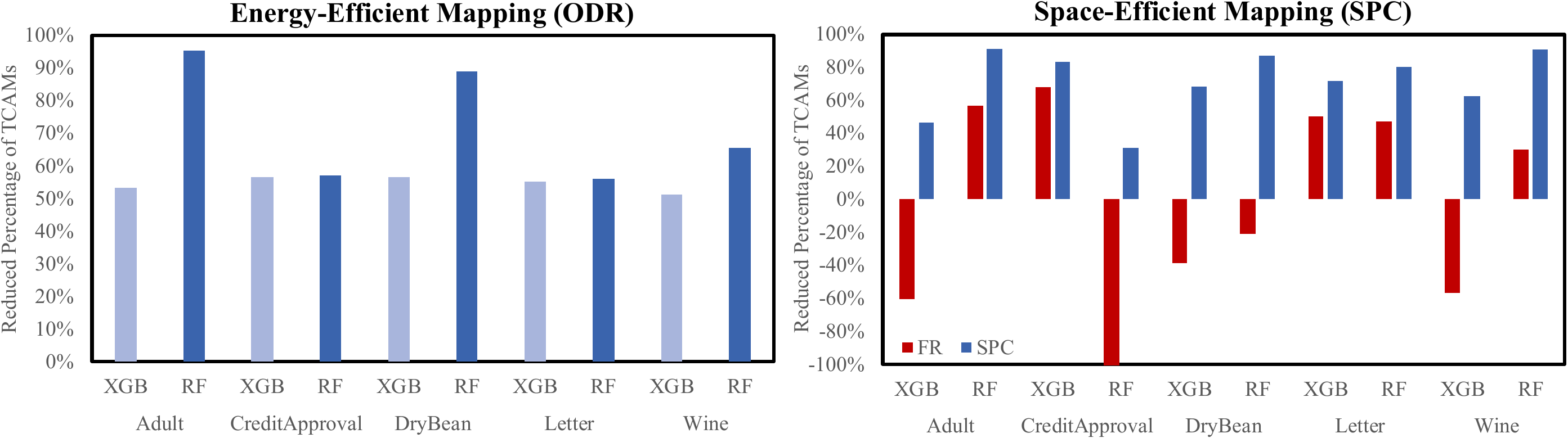}
  \caption{Comparison of different data placement strategies (XGB: XGBoost, RF: Random Forest).}
  \label{exp:fig_placement}
\end{figure*}

Noticing the extraordinary performance of RETENTION, we break down the framework and evaluate each component separately. This section analyzes the effectiveness of \textit{purity threshold pruning} by mapping Random Forest models with \textit{naive unified mapping}. By setting tolerance to \{1\%, 3\%, 5\%\}, \textit{purity threshold pruning} alone achieves considerable improvement, reducing CAM capacity requirement by 21.04\% to 99.93\% ($1.27\times$ to $1357.12\times$ improvement) compared to unpruned models. The relatively lower improvement on the Letter and Wine datasets stems from the increased number of classes, which makes it more challenging for nodes to achieve high purity. Pruning such nodes may result in significant accuracy degradation, which the tolerance mechanism prevents. Consequently, only a few higher-purity nodes can be pruned, thereby limiting the impact of \textit{purity threshold pruning}. We further scale tolerance up to \{10\%, 15\%, 20\%\} to investigate the correlation between tolerance and different datasets. As shown in the left part of Fig.~\ref{fig:exp_pruning}, the performance of \textit{purity threshold pruning} converges at different levels of tolerance, depending on task complexity. Though all can achieve over 90\% reduction, simple tasks easily converge with low tolerance (\{1\%, 3\%, 5\%\}, refer to the red columns), while the convergence for complex tasks occurs with relatively higher tolerance (\{10\%, 15\%, 20\%\}, refer to the blue columns). However, for imbalanced datasets such as Adult, higher tolerance may lead to a model always favoring the majority class. This should be carefully dealt with by leveraging techniques such as class weights, as discussed in Section~\ref{method:pruning}. The right part of Fig.~\ref{fig:exp_pruning} exhibits that the user-specified tolerance is highly correlated to testing accuracy loss. Since tolerance is the maximum expected accuracy degradation that is sacrificed to achieve lower model complexity, ideally the testing accuracy loss should be lower than or similar to tolerance. Results show that most of the testing accuracy losses are lower than the predefined tolerance, with only one case slightly higher, suggesting that \textit{purity threshold pruning} effectively reduces model complexity while ensuring controlled accuracy degradation. Models pruned with higher tolerance also follow this trend, though we omitted the statistics here for clarity.

\subsection{Experiment 3: Tree Mapping Scheme} \label{exp:placement}

After validating \textit{purity threshold pruning}, we estimate the effectiveness of the proposed data placement strategies. Fig.~\ref{exp:fig_placement} presents the results of implementing \textit{ODR} and \textit{SPC} on XGBoost and unpruned Random Forest. Since unpruned Random Forest is usually more complex than XGBoost, the improvement of applying \textit{ODR} and \textit{SPC} on Random Forest is often more obvious. For energy-efficient mapping, \textit{ODR} presents more than 50\% reduction in all cases. Since $\#unique\_conditions$ of the models trained on CreditApproval and Letter datasets is an order of magnitude fewer than in other datasets, the improvement observed in the Random Forests trained on these two datasets is relatively limited. For space-efficient mapping, \textit{SPC} exhibits notable improvement, with over 30\% reduction in every case, while \textit{FR} performs worse than the baseline in half of the cases. The underlying reason is that \textit{FR} only eliminates completely redundant path segments within a TCAM, whereas significant redundancy is still left behind. Thus, \textit{FR} beats \textit{naive independent mapping} only when the redundancy within a single tree is overwhelming. On the other hand, by greedily clustering similar paths together, \textit{SPC} minimizes redundancy while incurring less runtime computational overhead, consistently outperforming \textit{FR} and \textit{naive independent mapping} in all cases. Although applying \textit{SPC} to the Random Forest trained on the CreditApproval dataset yields minimum improvement, its performance is actually nearly optimal (\textit{SPC} requires 99 TCAMs, while the theoretical minimum is 98 for 6246 paths). The restricted improvement stems from the exceptional performance of \textit{naive independent mapping} on this model, which limits the potential for further improvement. Experimental results show that implementing the tree mapping scheme alone yields $1.46\times$ to $21.30\times$ capacity reduction, highlighting its effectiveness.

\subsection{Experiment 4: Number of Trees} \label{exp:tree}

\begin{figure*}[t]
  \centering
  \includegraphics[width=\linewidth, page=8]{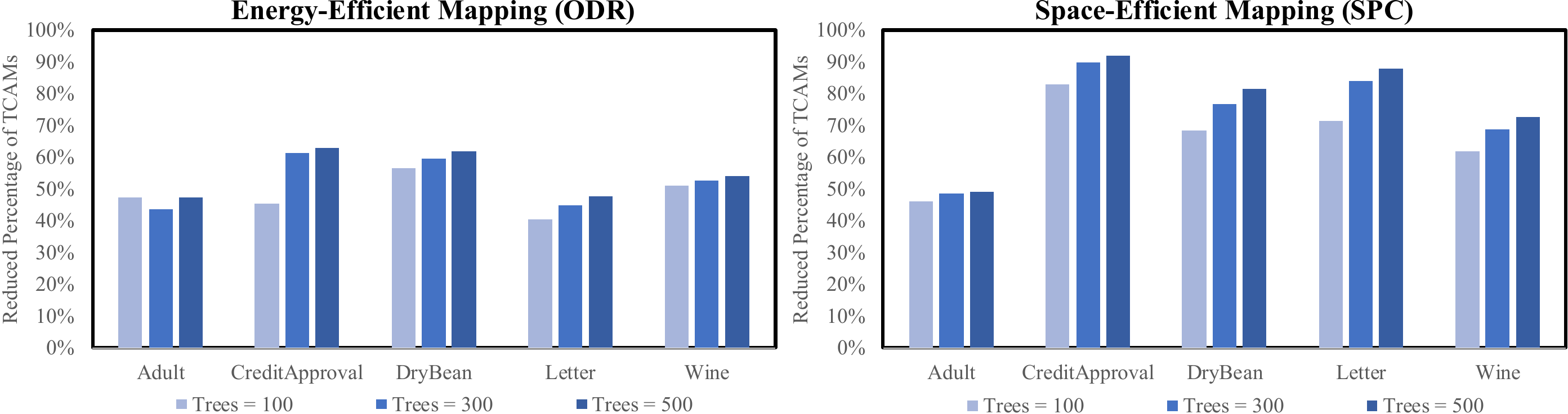}
  \caption{Impact of number of trees on the performance of data placement strategies.}
  \label{fig:exp_tree}
\end{figure*}
\begin{figure*}[t]
  \centering
  \includegraphics[width=\linewidth, page=9]{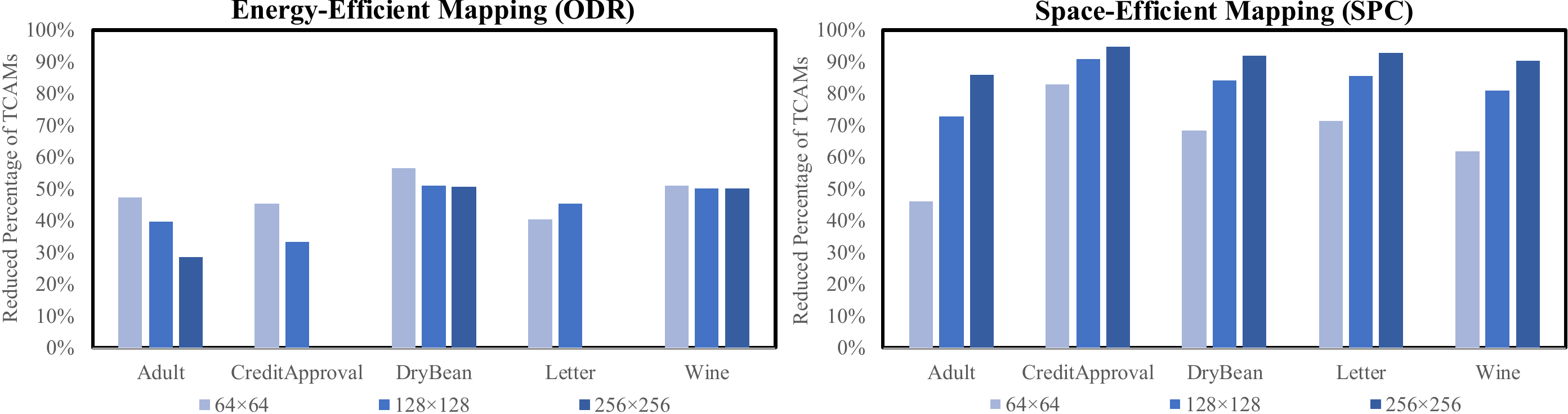}
  \caption{Impact of TCAM sizes on the performance of data placement strategies.}
  \label{fig:exp_TCAM}
\end{figure*}

Prior experiments demonstrate that RETENTION is extremely effective in minimizing CAM capacity requirement. In this section, we study the impact of number of trees on RETENTION's performance. Since XGBoost can achieve higher accuracy with an increased number of trees and appropriate hyperparameter tuning, XGBoost with \textit{num\_trees} set to \{100, 300, 500\} are evaluated in this experiment. As shown in Fig.~\ref{fig:exp_tree}, the performance of RETENTION tends to improve as \textit{num\_trees} increases because the inefficient issue is amplified by \textit{num\_trees}. However, since the training phase incorporates randomness, and the selected conditions for splitting nodes can strongly affect RETENTION's performance, the results may not strictly follow the trend.

\subsection{Experiment 5: Different TCAM Sizes} \label{exp:TCAM}

In addition to the number of trees, the size of the underlying TCAM is another factor that may affect RETENTION's performance. Fig.~\ref{fig:exp_TCAM} illustrates the impact of mapping XGBoost to TCAMs with three different sizes. Results suggest that as TCAM size increases, RETENTION yields gradually diminishing improvements in energy-efficient mapping, while offering notable gains in space-efficient mapping. Although the designs of both \textit{naive unified mapping} and \textit{ODR} are not directly correlated with TCAM size, it becomes increasingly difficult for \textit{ODR} to eliminate TCAMs as the size increases, since only those entirely filled with X states can be removed. In some edge cases, such as mapping the models trained on the CreditApproval or Letter dataset to $256 \times 256$ TCAMs, \textit{ODR} may offer no improvement because none of the TCAMs are fully redundant even after reordering. Conversely, increasing the TCAM size can sometimes create partially redundant TCAMs with only a few non-X-state cells. In such cases, \textit{ODR} can effectively mitigate this issue, allowing these TCAMs to be removed and lowering capacity requirement. For example, mapping the Letter model to $128 \times 128$ TCAMs may benefit from this effect. On the other hand, although improvement increases with TCAM size when RETENTION is applied for space-efficient mapping, CAM utilization actually decreases because larger TCAMs require more X states per row. However, since \textit{naive independent mapping} assigns each tree to a separate TCAM, larger capacities often fail to be fully exploited, leading to greater performance degradation compared to \textit{SPC}. Consequently, a smaller TCAM is recommended for RETENTION regardless of the optimization objective, though it must still be large enough to accommodate the longest path in the model when applying \textit{SPC}.

\begin{table*}[t]
\centering                 
\caption{detailed experimental statistics of RETENTION}
\label{tab:RETENTION}
\renewcommand{\arraystretch}{1.1}
\begin{tabular}{cc|cc|ccccc}    
\toprule                    
Model Type & Dataset & Tolerance & Accuracy Loss & Mapping & \#TCAMs (Mapping/Baseline) & Reduction & Required Capacity (KB)\\
\midrule

&  &  &  & ODR & 7156 / 1482120 & $207.12\times$ & 3578.0 \\
Random Forest & Adult & 3\% & 2.86\% & FR & 2313 / 41613 & $17.99\times$ & 1156.5 \\
&  &  &  & SPC & 495 / 41613 & $84.07\times$ & 247.5 \\
\midrule

&  &  &  & ODR & 33 / 3038 & $92.06\times$ & 16.5 \\
Random Forest & CreditApproval & 3\% & 1.14\% & FR & 26 / 144 & $5.54\times$ & 13.0 \\
&  &  &  & SPC & 12 / 144 & $12.00\times$ & 6.0 \\
\midrule

&  &  &  & ODR & 2833 / 537419 & $189.70\times$ & 1416.5 \\
Random Forest & DryBean & 3\% & 2.76\% & FR & 824 / 7693 & $9.34\times$ & 412.0 \\
&  &  &  & SPC & 113 / 7693 & $68.08\times$ & 56.5 \\
\midrule

&  &  &  & ODR & 4783 / 20825 & $4.35\times$ & 2391.5 \\
Random Forest & Letter & 3\% & 2.26\% & FR & 4278 / 15120 & $3.53\times$ & 2139.0 \\
&  &  &  & SPC & 1641 / 15120 & $9.21\times$ & 820.5 \\
\midrule

&  &  &  & ODR & 28322 / 144824 & $5.11\times$ & 14161.0 \\
Random Forest & Wine & 3\% & 0.23\% & FR & 8424 / 21608 & $2.57\times$ & 4212.0 \\
&  &  &  & SPC & 1163 / 21608 & $18.58\times$ & 581.5 \\
\midrule

&  &  &  & ODR & 227 / 486 & $2.14\times$ & 113.5 \\
XGBoost & Adult & - & 0.00\% & FR & 162 / 101 & Failed & 81.0 \\
&  &  &  & SPC & 54 / 101 & $1.87\times$ & 27.0 \\
\midrule

&  &  &  & ODR & 37 / 85 & $2.30\times$ & 18.5 \\
XGBoost & CreditApproval & - & 0.00\% & FR & 32 / 100 & $3.13\times$ & 16.0 \\
&  &  &  & SPC & 17 / 100 & $5.88\times$ & 8.5 \\
\midrule

&  &  &  & ODR & 3382 / 7790 & $2.30\times$ & 1691.0 \\
XGBoost & DryBean & - & 0.00\% & FR & 908 / 654 & Failed & 454.0 \\
&  &  &  & SPC & 206 / 654 & $3.17\times$ & 103.0 \\
\midrule

&  &  &  & ODR & 1321 / 2960 & $2.24\times$ & 660.5 \\
XGBoost & Letter & - & 0.00\% & FR & 1292 / 2600 & $2.06\times$ & 646.0 \\
&  &  &  & SPC & 740 / 2600 & $3.51\times$ & 370.0 \\
\midrule

 &  &  &  & ODR & 2722 / 5565 & $2.04\times$ & 1361.0 \\
XGBoost & Wine & - & 0.00\% & FR & 1117 / 711 & Failed & 558.5 \\
&  &  &  & SPC & 267 / 711 & $2.66\times$ & 133.5 \\

\bottomrule
\end{tabular}
\end{table*}

\begin{table}[t]
\centering
\caption{energy consumption and latency}
    \centering
    \renewcommand{\arraystretch}{1.1}
    \begin{tabular}{c|cc}
        \toprule
        Operation (Per TCAM) & Energy & Latency \\
        \midrule
        Query Packing (Random) & $6.27 \mu J$ & $14.99 ns$ \\
        Query Packing (Sequential) & $5.34 \mu J$  & $14.70 ns$\\
        TCAM Search \cite{ReRAMTCAM}& $2.09 pJ$ & $0.96 ns$ \\
        \bottomrule
    \end{tabular}
    \label{tab:energy}
\end{table}

\subsection{Experiment 6: Energy Consumption and Latency} \label{exp:energy}
The workflow of CAM-based inference is (1) encode input features into a binary sequence, (2) pack and distribute queries to CAMs, (3) search in CAMs, and (4) generate a prediction based on the search results. Since (1) and (4) are mapping-independent, here we only discuss the energy consumption and latency for (2) and (3). Table~\ref{tab:energy} shows the detailed statistics for query packing and TCAM search with size $64\times64$, measured with the largest $\#unique\_conditions$  (i.e., 5848) observed from the models listed on Table~\ref{tab:RETENTION}. As the thresholds for each feature are sorted for fast encoding, mapping with clustering or column reordering (i.e., \textit{ODR}, \textit{SPC}, \textit{naive independent mapping}) requires random accessing the encoded sequence during query packing, while \textit{naive unified mapping} accesses sequentially. However, even the longest encoded sequence occupies only 5.71KB, and therefore it can be perfectly cached with minimal influence on latency.

As discussed in Section~\ref{method:mapping}, \textit{naive unified mapping} and \textit{ODR} both require $\left\lceil \frac{\#\textit{unique\_conditions}}{S} \right\rceil$ times of query packing given the size of CAM $S \times S$. Therefore, 
\textit{ODR} substantially reduces capacity requirement at the cost of slightly higher energy consumption and latency than \textit{naive unified mapping}, as the savings from reducing search operations is negligible compared to the cost of random accessing. On the other hand, since both \textit{SPC} and \textit{naive independent mapping} require \#TCAM times of query packing with random accessing, \textit{SPC} is more efficient than \textit{naive independent mapping} considering capacity requirement, energy consumption, and latency.

\section{Conclusion} \label{sec:conclusion}
nvTCAM shows great promise to accelerate tree-based model inference effectively and efficiently, yet the capacity requirement of CAM is unacceptably high, and most of the cells are storing the X state for format alignment, which is redundant. In this work, we introduce RETENTION, an end-to-end framework designed to reduce CAM capacity requirement for tree-based model inference. RETENTION introduces \textit{purity threshold pruning} to minimize model complexity while ensuring controlled accuracy degradation for bagging-based models. A tree mapping scheme with two data placement strategies, \textit{occurrence-based double reordering} and \textit{similarity-based path clustering}, is proposed to further alleviate memory redundancy for energy-efficient mapping and space-efficient mapping. Experimental results show that implementing the tree mapping scheme alone achieves $1.46\times$ to $21.30\times$ capacity reduction, while the full RETENTION framework yields $4.35\times$ to $207.12\times$ improvement with less than 3\% accuracy loss. These results demonstrate that RETENTION is extremely effective in reducing CAM capacity requirement, offering a resource-efficient solution for tree-based model acceleration. By implementing RETENTION, resource retention becomes feasible, bringing CAM-based acceleration closer to practicality in resource-constrained environments.

\section*{Acknowledgments}
The authors acknowledge the use of ChatGPT and Gemini for language editing and text refinement . All technical content, experimental design, data analysis, and scientific conclusions were solely developed and verified by the authors.

\bibliographystyle{IEEEtran}
\bibliography{references}

\vspace{-30px}

\begin{IEEEbiography}[{\includegraphics[width=1in,height=1.25in,clip,keepaspectratio]{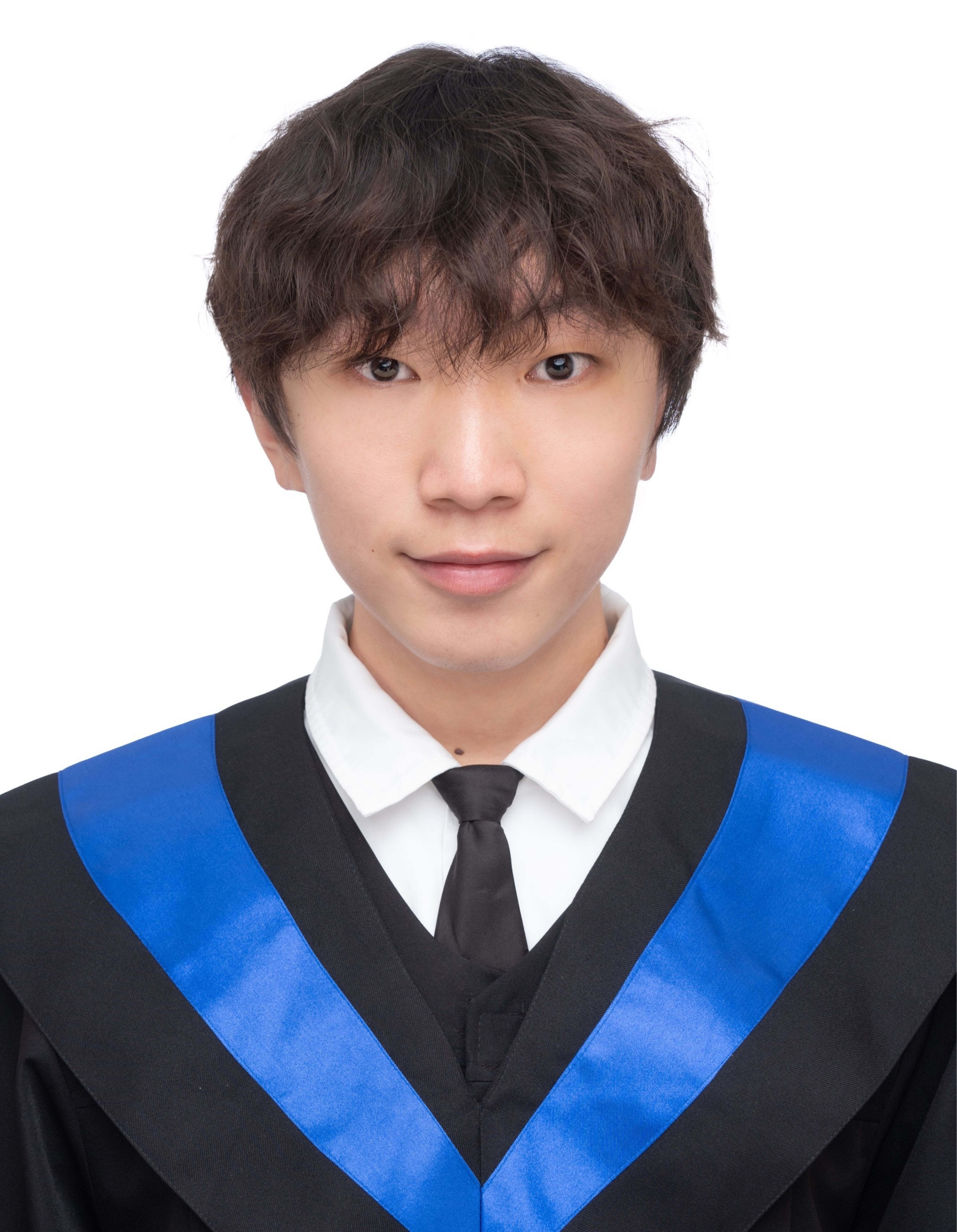}}]{Yi-Chun Liao}
received his B.Sc. degree from the Department of Computer Science and Information Engineering, National Taiwan University, Taipei, Taiwan, in 2026. He is currently pursuing the Ph.D. degree with the Department of Electrical and Computer Engineering, Duke University, Durham, NC, USA. His primary research interests include in-memory computing, machine learning, and algorithm-hardware co-design.
\end{IEEEbiography}

\vspace{-30px}

\begin{IEEEbiography}[{\includegraphics[width=1in,height=1.25in,clip,keepaspectratio]{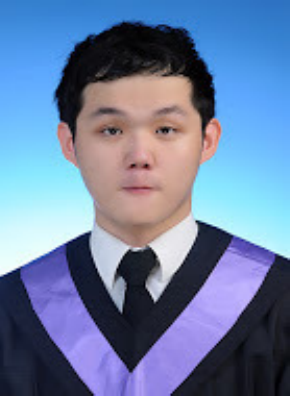}}]{Chieh-Lin Tsai}
received his M.S. degree from the Department of Computer Science, National Tsing-Hua University, Hsinchu, Taiwan, in 2017. He is currently
a Ph.D. candidate of the Department
of Computer Science and Information
Engineering, National Taiwan University. His primary research interests include non-volatile memory systems and system-level design with in-memory processing.
\end{IEEEbiography}

\vspace{-30px}

\begin{IEEEbiography}[{\includegraphics[width=1in,height=1.25in,clip,keepaspectratio]{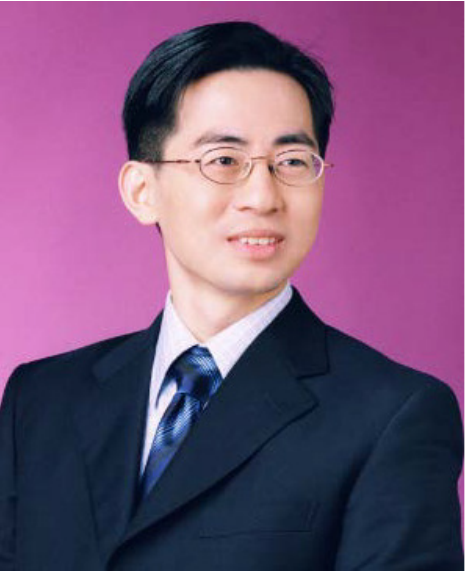}}]{Yuan-Hao Chang} 
received his Ph.D. degree from the Department of Computer Science and Information Engineering, National Taiwan University, Taipei, Taiwan. He is currently a Professor of Department of Computer Science and Information Engineering, National Taiwan University.  His primary research interests include memory/storage systems, operating systems, embedded systems, and real-time systems. He is an IEEE Fellow.
\end{IEEEbiography}

\vspace{-30px}

\begin{IEEEbiography}
[{\includegraphics[width=1in,height=1.25in,clip,keepaspectratio]{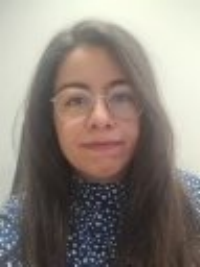}}]{Camélia Slimani} 
received her Ph.D. degree in Computer Science from the University of Western Brittany, France, in 2022. She was then a postdoctoral fellow at ENSTA Bretagne, France. She is currently an Associate Professor at the National Polytechnic Institute of Toulouse, France, and a member of the IRIT laboratory. Her primary research interests include optimizing machine learning algorithms for memory-constrained environments.
\end{IEEEbiography}

\vspace{-30px}

\begin{IEEEbiography}[{\includegraphics[width=1in,height=1.25in,clip,keepaspectratio]{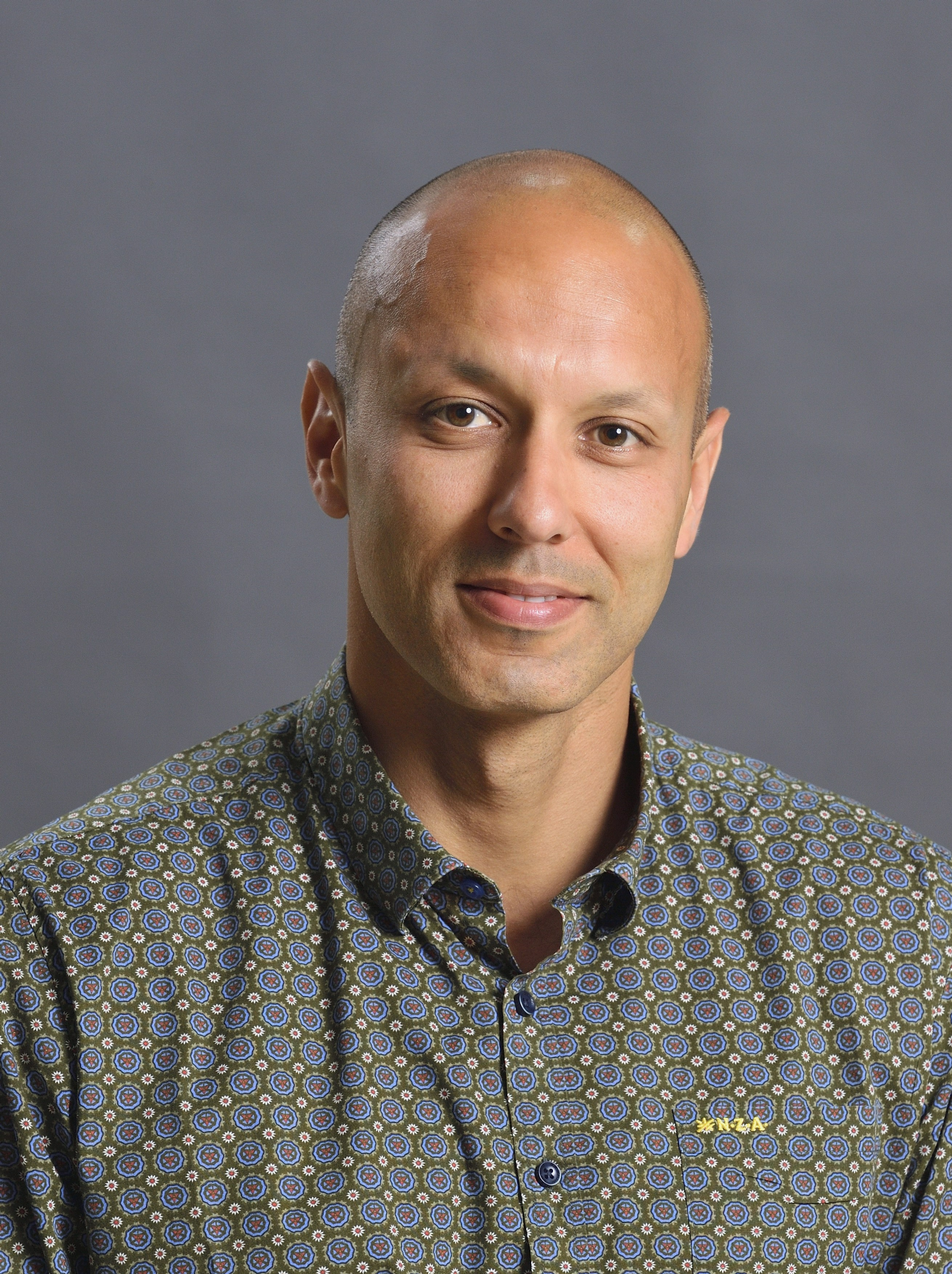}}]{Jalil Boukhobza} 
received his Electrical Engineering (with Hons.) degree from INELEC, Boumerdes, Algeria, in 1999, and the M.Sc. and Ph.D. degrees in Computer Science from the University of Versailles, France, in 2000 and 2004, respectively. He is currently a Professor of the ENSTA, Institut Polytechnique de Paris, and a member of Lab-STICC. His primary research interests include storage system design, performance evaluation and energy optimization, and operating system design. He is an IEEE Senior member.
\end{IEEEbiography}

\vspace{-30px}

\begin{IEEEbiography}[{\includegraphics[width=1in,height=1.25in,clip,keepaspectratio]{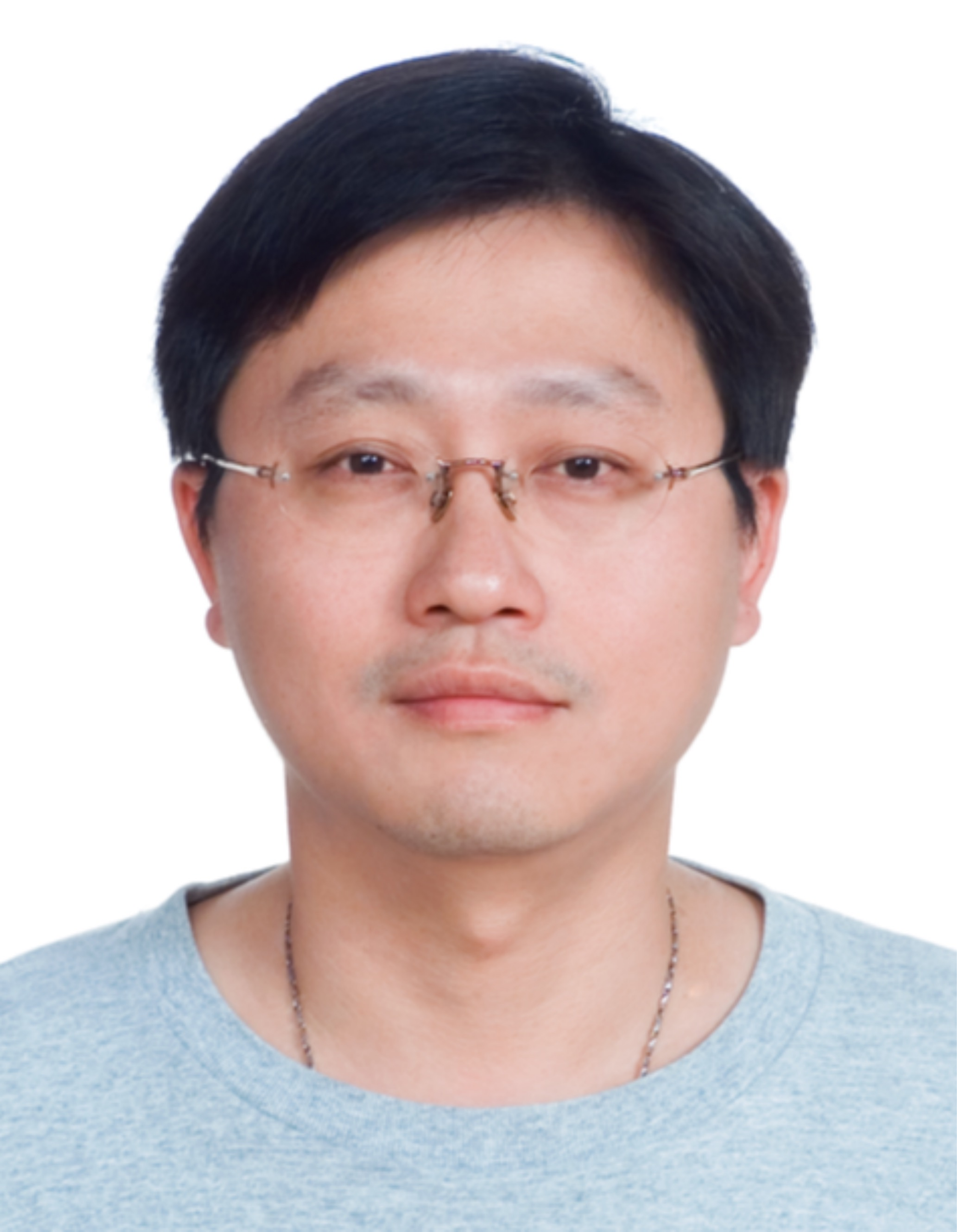}}]{Tei-Wei Kuo} 
received his B.S.E. and Ph.D. degrees in Computer Science from National Taiwan University and University of Texas at Austin in 1986 and 1994, respectively. He is the CTO of Delta Electronics (2024.02-now). His primary research interests include embedded systems, non-volatile memory software designs, neuromorphic computing, and real-time systems. He is a Fellow of ACM, IEEE, and US National Academy of Inventors. He is the Chair of ACM SIGAPP (since 2023).
\end{IEEEbiography}

\end{document}